\documentclass{article} 
\usepackage{nips12submit_e,times}
\usepackage{graphicx} 
\usepackage{amsmath}
\usepackage{amssymb}
\usepackage{subfigure} 
\usepackage[numbers]{natbib} 
\usepackage{algorithm,algorithmic}
\usepackage{caption}
\usepackage{mathtools}

\title{Metric-Free Natural Gradient for Joint-Training of Boltzmann Machines}

\author{
Guillaume Desjardins, Razvan Pascanu, Aaron Courville and Yoshua Bengio \\
D\'{e}partement d'informatique et de recherche op\'{e}rationnelle \\
Universit\'{e} de Montr\'{e}al \\
}

\newcommand{\h}{h^{(1)}}
\newcommand{\hh}{h^{(2)}}

\nipsfinalcopy 

\begin{document}

\maketitle

\begin{abstract}
This paper introduces the Metric-Free Natural Gradient (MFNG) algorithm for
training Boltzmann Machines. Similar in spirit to the Hessian-Free
method of \citet{martens2010hessian}, our algorithm belongs to the family
of truncated Newton methods and exploits an efficient matrix-vector product
to avoid explicitly storing the natural gradient metric $L$. This metric
is shown to be the expected second derivative of the log-partition function
(under the model distribution), or equivalently, the covariance of the vector of
partial derivatives of the energy function. We evaluate our method on the
task of joint-training a 3-layer Deep Boltzmann Machine and show that
MFNG does indeed have faster per-epoch convergence compared to Stochastic
Maximum Likelihood with centering, though wall-clock performance is
currently not competitive.
\end{abstract}

\section{Introduction}

Boltzmann Machines (BM) have become a popular method in Deep Learning for
performing feature extraction and probability modeling. The emergence of these
models as practical learning algorithms stems from the development of efficient
training algorithms, which estimate the negative log-likelihood gradient by either
contrastive \cite{Perpinan+Hinton-2005-small} or stochastic \cite{Tieleman08,Younes1999}
approximations. However, the success of these models has for the most part been limited to
the Restricted Boltzmann Machine (RBM) \cite{Freund+Haussler92}, whose architecture allows for
efficient exact inference. Unfortunately, this comes at the cost of the model's
representational capacity, which is limited to a single layer of latent
variables. The Deep Boltzmann Machine (DBM) \cite{Salakhutdinov2009}
addresses this by defining a joint energy function over multiple disjoint layers of
latent variables, where interactions within a layer are prohibited. While this affords the model a rich inference scheme
incorporating top-down feedback, it also makes training much more difficult,
requiring until recently an initial greedy layer-wise pretraining
scheme. Since, \citet{Montavon2012} have shown that this difficulty stems from
an ill-conditioning of the Hessian matrix, which can be addressed by a simple
reparameterization of the DBM energy function, a trick called {\it centering} 
(an analogue to centering and skip-connections found in the deterministic neural network
literature \cite{Schraudolph-1998,Raiko-2012-small}).
As the barrier to joint-training
\footnote{Joint-training refers to the act of jointly optimizing $\theta$ (the
concatenation of all model parameters, across all layers of the DBM) through
maximum likelihood. This is in contrast to \cite{Salakhutdinov2009}, where
joint-training is preceded by a greedy layer-wise pretraining strategy.}
is overcoming a challenging optimization problem, it is apparent that
second-order gradient methods might prove to be more effective than simple
stochastic gradient methods. This should prove especially important as 
we consider models with increasingly complex posteriors or higher-order
interactions between latent variables.

To this end, we explore the use of the Natural Gradient \cite{amari98natural},
which seems ideally suited to the stochastic
nature of Boltzmann Machines. Our paper is structured as follows.
Section~\ref{sec:background} provides a detailed derivation of the natural
gradient, including its specific form for BMs. While most of these
equations have previously appeared in \cite{Amari-1992}, our derivation aims
to be more accessible as it attempts to derive the natural gradient from basic
principles, while minimizing references to Information Geometry.
Section~\ref{sec:algo} represents the true contribution of the paper: a
practical natural gradient algorithm for BMs which exploits the persistent
Markov chains of Stochastic Maximum Likelihood (SML) \cite{Tieleman08}, with a
Hessian-Free (HF) like algorithm \cite{martens2010hessian}. The method, named
Metric-Free Natural Gradient (MFNG) (in recognition of the similarities of our
method to HF), avoids explicitly storing the natural gradient metric $L$ and
uses a linear solver to perform the required matrix-vector product
$L^{-1} \mathbb{E}_q\left[\nabla \log p_\theta\right]$.
Preliminary experimental results on DBMs are presented in
Section~\ref{sec:experiments}, with the discussion appearing in
Section~\ref{sec:discussion}.

\section{The Natural Gradient}
\label{sec:background}

\subsection{Motivation and Derivation}
\label{sec:motivation}

The main insight behind the natural gradient is that the space of all
probability distributions $\mathcal{P} = \{p_\theta(x);\ \theta \in \Theta, x
\in \chi \}$ forms a Riemannian manifold. Learning, which typically proceeds by
iteratively adapting the parameters $\theta$ to fit an empirical distribution
$q$, thus traces out a path along this manifold. An immediate consequence is
that following the direction of steepest descent in the original Euclidean
parameter space does not correspond to the direction of
steepest descent along $\mathcal{P}$. To do so, one needs to account for the
metric describing the local geometry of the manifold, which is given by the Fisher Information
matrix \cite{Amari-1985}, shown in Equation~\ref{eq:natural_fisher}. While this metric is
typically derived from Information Geometry, a derivation more
accessible to a machine learning audience can be obtained as follows.

The natural gradient aims to find the search direction $\Delta\theta$ which
minimizes a given objective function, such that the Kullback--Leibler divergence
$KL(p_{\theta} \parallel p_{\theta+\Delta\theta})$ remains constant throughout
optimization. This constraint ensures that we make constant progress
regardless of the curvature of the manifold $\mathcal{P}$ and enforces an
\emph{invariance to the parameterization of the model}.  The natural gradient
for maximum likelihood can thus be formalized as:
\begin{eqnarray}
\label{eq:natural_objective}
\nabla_N := \Delta\theta^* & \leftarrow & 
    \text{argmin}_{\Delta\theta}\ 
    \mathbb{E}_{q} \left[ - \log p_{\theta+\Delta\theta}(x) \right]\\
 & \text{s.t.} & KL(p_{\theta} \parallel p_{\theta+\Delta\theta}) = \mathrm{const.} \nonumber
\end{eqnarray}

In order to derive a useful parameter update rule, we will consider the KL
divergence under the assumption $\Delta\theta \rightarrow 0$. We also assume we
have a discrete and bounded domain $\chi$ over which we define the probability
mass function\footnote{When clear from context, we will drop the argument of
$p_\theta$ to save space.} $p_\theta$.  Taking the Taylor series expansion of
$\log p_{\theta+\Delta\theta}$ around $\theta$, and denoting $\nabla f$ as the
column vector of partial derivatives with $\frac{\partial f}{\partial \theta_i}$ as
the $i$-th entry, and $\nabla^2 f$ the Hessian matrix with $\frac{\partial^2
f}{\partial \theta_i \partial \theta_j}$ in position $(i,j)$, we have:
\begin{eqnarray}
    \label{eq:KL}
KL(p_{\theta} \parallel p_{\theta+\Delta\theta}) & \approx &
    \sum_{\chi}p_\theta \log p_\theta
    -\sum_{\chi}p_{\theta}
    \left[
        \log p_{\theta} 
        + \left(\nabla \log p_{\theta}\right)^T \Delta\theta
        + \frac{1}{2} \Delta\theta^T \left( \nabla^2 \log p_{\theta} \right) \Delta\theta
    \right] \nonumber \\
    \label{eq:expected_hessian}
    & = & \frac{1}{2}\Delta\theta^T
    \mathbb{E}_{p_\theta}
        \left[ - \nabla^2 \log p_\theta \right]
        \Delta\theta
\end{eqnarray}

with the transition stemming from the fact that
$\sum_\chi p_\theta \frac{\partial \log p_\theta}{\partial \theta_i} =
\frac{\partial}{\partial \theta_i} \sum_{x \in \chi} p_\theta(x) = 0$.
Replacing the objective function of Equation~\ref{eq:natural_objective} by its
first-order Taylor expansion and rewriting the constraint as a Lagrangian, we
arrive at the following formulation for $\mathcal{L}(\theta,
\Delta\theta)$, the loss function which the natural gradient seeks to minimize.
\begin{eqnarray*}
\mathcal{L}(\theta,\Delta\theta) & = &
    \mathbb{E}_{q}\left[ -\log p_\theta \right] + 
    \mathbb{E}_{q}\left[ - \nabla \log p_\theta \right]^T \Delta\theta
    +\frac{\lambda}{2}\Delta\theta^T
    \mathbb{E}_{p_\theta}
    \left[ -\nabla^2 \log p_\theta \right]
    \Delta\theta.
\end{eqnarray*}

Setting $\frac{\partial\mathcal{L}}{\partial\Delta\theta}$ to zero yields the natural gradient direction $\nabla_N$:
\begin{eqnarray}
    \label{eq:natural_hessian}
    \nabla_N = L^{-1} \mathbb{E}_q \left[ \nabla \log p_\theta \right] \text{ with }
    L & = & \mathbb{E}_{p_\theta}
        \left[ - \nabla^2 \log p_\theta \right] \\
    \label{eq:natural_fisher}
    \text{ or equivalently } L & = & \mathbb{E}_{p_\theta}
    \left[
        \nabla \log p_\theta
        \nabla^T \log p_\theta
    \right] 
\end{eqnarray}
While its form is reminiscent of the Newton direction, the natural gradient
multiplies the estimated gradient by the inverse of the expected Hessian of
$\log p_\theta$ (Equation~\ref{eq:natural_hessian}) or equivalently by the
Fisher Information matrix (FIM, Equation~\ref{eq:natural_fisher}).  The
equivalence between both expressions can be shown trivially, with the details
appearing in the Appendix. We stress that both of these expectations are
computed with respect to the {\it model distribution}, and thus computing the
metric $L$ does not involve the empirical distribution in any way.  The FIM for
Boltzmann Machines is thus {\it not equal} to the uncentered covariance of the
maximum likelihood gradients. In the following, we pursue our derivation from
the form given in Equation~\ref{eq:natural_fisher}.

\subsection{Natural Gradient for Boltzmann Machines}
\label{sec:natural_boltzmann}

\paragraph{Derivation.}
Boltzmann machines define a joint distribution over a vector of binary random
variables $x \in \{0,1\}^N$ by way of an energy function $E(x) = - \sum_{k < l}
W_{kl} x_k x_l -\sum_k b_k x_k$, with weight matrix $W\in \mathbb{R}^{N\times
N}$ and bias vector $b \in \mathbb{R}^N$. Energy and probability are related by
the Boltzmann distribution, such that $p(x) = \frac{1}{Z}
\exp\left(-E(x)\right)$, with $Z$ the partition function defined by
$Z = \sum_x \exp \left(-E(x)\right)$.

Starting from the expression of $L$ found in Equation~\ref{eq:natural_hessian},
we can derive the natural gradient metric for Boltzmann Machines.
\begin{eqnarray*}
    L^{(BM)} & = & 
    \mathbb{E}_{p_\theta} \left[ \nabla^2 E(x) + \nabla^2 \log Z \right] =
    \mathbb{E}_{p_\theta} \left[ \nabla^2 \log Z \right]\\
\end{eqnarray*}

The natural gradient metric for first-order BMs takes on a surprisingly simple form: it is
the expected Hessian of the log-partition function. With a few lines
of algebra (whose details are presented in the Appendix), we can rewrite it as
follows:
\begin{eqnarray}
    \label{eq:dbm_metric}
    L^{(BM)} & = &
    \mathbb{E}_{p_\theta}  \left[
        \left( \nabla E(x) - \mathbb{E}_{p_\theta}\left[ \nabla E(x) \right] \right)^T
        \left( \nabla E(x) - \mathbb{E}_{p_\theta}\left[ \nabla E(x) \right] \right)
        \right]
\end{eqnarray}

$L^{(BM)}$ is thus given by the covariance of $\nabla E$, measured under the
model distribution $p_\theta$.  Concretely, if we denote $W_{kl}$ and $W_{mn}$
as the i and j-th parameters of the model respectively, the entry $L_{ij}$ will
take on the value $- \mathbb{E} \left[ x_k x_l x_m x_n \right] + \mathbb{E}
\left[ x_k x_l \right] \mathbb{E} \left[ x_m x_n \right]$.

\paragraph{Discussion.}
When computing the Taylor expansion of the KL divergence in
Equation~\ref{eq:KL}, we glossed over an important detail. Namely, how to
handle latent variables in $p_\theta(x)$, a topic first discussed in
\cite{Amari-1992}. If $x = [v,h]$, we could just as easily have derived the natural
gradient by considering the constraint $KL\left(\sum_h p_\theta(v,h) \parallel
\sum_h p_{\theta + \Delta\theta}(v,h) \right) = \text{const}$. Alternatively,
since the distinction between visible and hidden units is entirely artificial
(since the KL divergence does not involve the empirical distribution),
we may simply wish to consider the distribution obtained by analytically
integrating out a maximal number of random variables.  In a DBM, this would
entail marginalizing over all odd or even layers, a strategy employed with great
success in the context of AIS \cite{Salakhutdinov2009}.  In this work however, we only
consider the metric obtained by considering the $KL$ divergence between the
full joint distributions $p_\theta$ and $p_{\theta + \Delta\theta}$.

\section{Metric-Free Natural Gradient Implementation}
\label{sec:algo}

We can compute the natural gradient $\nabla_N$ by first replacing the expectations of
Equation~\ref{eq:dbm_metric} by a finite sample approximation. We can do
this efficiently by reusing the model samples generated by the persistent
Markov chains of SML. Given the size of the matrix being estimated however, we
expect this method to require a larger number of chains than is typically used.
The rest of the method is similar to the Hessian-Free (HF) algorithm of
\citet{martens2010hessian}: we exploit an efficient matrix-vector
implementation combined with a linear-solver, such as Conjugate Gradient or
MinRes\cite{paige:1975}, to solve the system
$Ly = \mathbb{E}_q \left[ \nabla \log p_\theta \right]$
for $y \in \mathbb{R}^N$. Additionally, we replace the expectation on the
rhs. of this previous equation by an average computed over a mini-batch of
training examples (sampled from the empirical distribution $q$), as is
typically done in the stochastic learning setting.

For Boltzmann Machines, the matrix-vector product $Ly$ can be computed
in a straightforward manner, without recourse to Pearlmutter's R-operator
\cite{Pearlmutter-1994}.  Starting from a sampling approximation to
Equation~\ref{eq:dbm_metric}, we simply push the dot product inside of the
expectation as follows:

\begin{eqnarray}
    \label{eq:matrix_vector}
    L^{(BM)}y & \approx & \left(S - \bar{S}\right)^T \left[ \left(S - \bar{S}\right) y \right] \\
    \text{ with } & &
        S \in \mathbb{R}^{M \times N} \text{, the matrix with entries } 
        s_{mj} = \frac {\partial E(x_m)}{\partial \theta_j} \nonumber \\
    \text{ and } &  &
        \bar{S} \in \mathbb{R}^N \text{, the vector with entries } 
        s_j = \frac{1}{M} \sum_m s_{mj} \nonumber \\
    \text{ and } &  &
        x_m \sim p_\theta(x), m \in [1,M]. \nonumber
\end{eqnarray}

By first computing the matrix-vector product $(S - \bar{S}) y$, we can easily
avoid computing the full $N \times N$ matrix $L$. Indeed, the result of this
operation is a vector of length $M$, which is then left-multiplied by a matrix
of dimension $N \times M$, yielding the matrix-vector product $Ly \in
\mathbb{R}^N$. A single iteration of the MFNG is presented in
Algorithm~\ref{alg:MFNG}. A full open-source implementation is also available
online.\footnote{https://github.com/gdesjardins/MFNG}.

\begin{algorithm}[t]
    \caption{
    ${\tt MFNG\_iteration}(\theta, \mathcal{X^+}, \mathcal{Z}^-_{old})$ \vspace{1mm}\\
    $\theta$: parameters of the model. $N \vcentcolon= \mid \theta \mid$. \vspace{1mm}\\
    $\mathcal{X}^+$: mini-batch of training examples, with $\mathcal{X}^+ = \{x_m; m \in [1,M]\}$. \vspace{1mm}\\
    $\mathcal{Z}_{old}^-$: previous state of Markov chains, with
        $\mathcal{Z} = \{z_m \vcentcolon = (v_m, \h_m, \hh_m); m \in [1,M]\}$\\
    }
\label{alg:MFNG}
\begin{algorithmic}
\vspace{2mm}
\STATE $\bullet$ Generate positive phase samples
       $\mathcal{Z}^+ = \{z^+_m \vcentcolon = (x_m, h^{(1)+}_m, h^{(2)+}_m); m \in [1,M]\}$
\vspace{1mm} 
\STATE $\bullet$ Initializing $M$ Markov chains from state $\mathcal{Z}^-_{old}$,
       generate negative phase samples $\mathcal{Z}^-_{new}$.
\vspace{1mm}
\STATE $\bullet$ Compute the vectors
$s^+_m = \frac{\partial E(z^+_m)}{\partial \theta}$ and
$s^-_m = \frac{\partial E(z^-_m)}{\partial \theta}$, $\forall m$.
\vspace{1mm}
\STATE $\bullet$ Compute negative log-likelihood gradient as
       $g = \frac{1}{M} \sum_m \left(s^+_m - s^-_m\right)$.
\vspace{1mm}
\STATE $\bullet$ Denote $S \in \mathbb{R}^{M \times N}$
as the matrix with rows $s^-_m$ and $\bar{S} = \frac{1}{M} \sum_m s^-_m$.
\vspace{2mm}
\STATE {\# \it Solve the system ``$Ly=g$'' for y, given $L=(S - \bar{S})^T (S - \bar{S})$ and an initial zero vector.}
\STATE {\# \it computeLy is a function which performs equation ~\ref{eq:matrix_vector}, without instantiating $L$.}
\vspace{1mm}
\STATE $\bullet$ $\nabla_N \theta \leftarrow {\tt CGSolve}(
    {\tt computeLy}, S, g, {\tt zeros}(N))$
\end{algorithmic}
\end{algorithm}

\section{Experiments}
\label{sec:experiments}

We performed a proof-of-concept experiment to determine
whether our Metric-Free Natural Gradient (MFNG) algorithm is suitable for
joint-training of complex Boltzmann Machines. To this end, we compared our
method to Stochastic Maximum Likelihood and a diagonal approximation of MFNG on
a 3-layer Deep Boltzmann Machine trained on MNIST~\cite{LeCun98-small}. All
algorithms were run in conjunction with the centering strategy of
\citet{Montavon2012}, which proved crucial to successfully joint-train all
layers of the DBM (even when using MFNG)
\footnote{The centering coefficients were initialized as in
\cite{Montavon2012}, but were otherwise held fixed during training.}.
We chose a small 3-layer DBM
with 784-400-100 units at the first, second and third layers respectively, to
be comparable to \cite{Montavon2012arxiv}. Hyper-parameters were varied as follows.
For inference, we ran $5$ iterations of either mean-field as implemented in
\cite{Salakhutdinov2009} or Gibbs sampling. The learning rate was kept
fixed during training and chosen from the set $\{5\cdot10^{-3},
10^{-3}, 10^{-4}\}$. For MinRes, we set the damping coefficient to $0.1$
and used a fixed tolerance of $10^{-5}$ (used to determine convergence).
Finally, we tested all algorithms on minibatch sizes of either $25$, $128$ or
$256$ elements \footnote{We expect larger minibatch sizes to be preferable,
however simulating this number of Markov chains in parallel (on top of all
other memory requirements) was sufficient to hit the memory bottlenecks of
GPUs.}.  Finally, since we are comparing optimization algorithms,
hyper-parameters were chosen based on the training set likelihood (though we
still report the associated test errors). All experiments used the MinRes
linear solver, both for its speed and its ability to return pseudo-inverses
when faced with ill-conditioning.

\begin{figure}
    \centering
    \hspace*{-0.5cm}
    \subfigure
    {
        \label{fig:results_epoch}
        \includegraphics[scale=0.38]{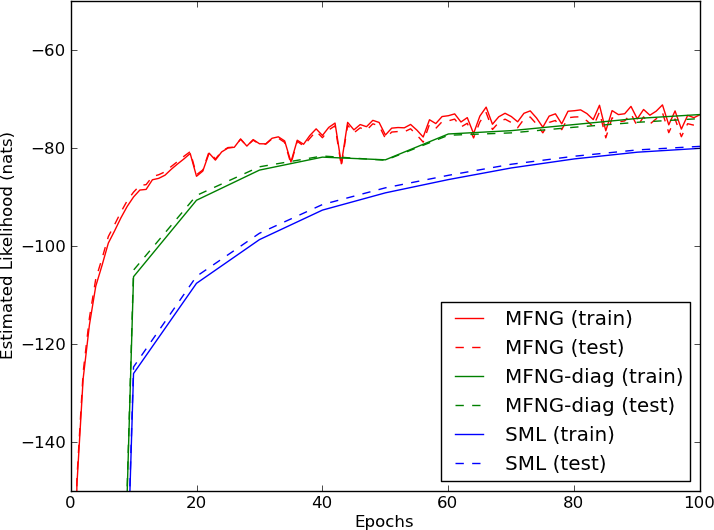}
    }
    \subfigure
    {
        \label{fig:results_cputime}
        \includegraphics[scale=0.38]{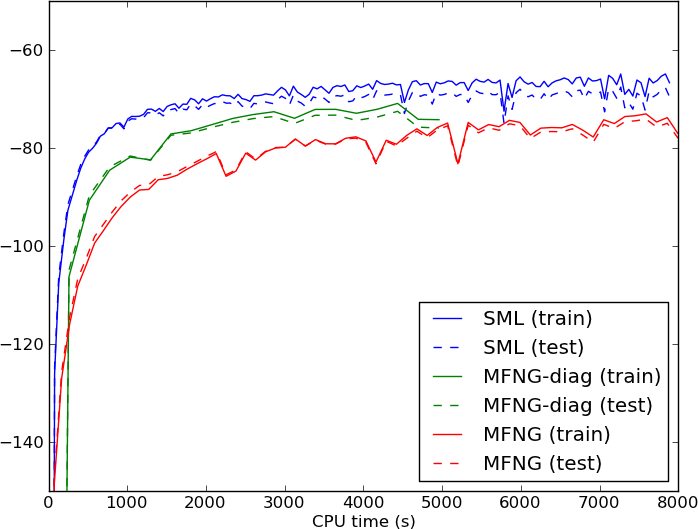}
    }
    \caption[]{Estimated model likelihood as a function of (left) epochs and
    (right) CPU-time for MFNG, its diagonal approximation (MFNG-diag) and SML.
    All methods were run in conjunction with the DBM centering trick
    \cite{Montavon2012}, with centering coefficients held fixed during
    training. Our grid search yielded the following hyper-parameters: batch
    size of $256/128$ for MFNG(-diag)/SGD; 5 steps of mean-field /
    sampling-based inference for MFNG(-diag)/SGD and a learning rate of
    $5\cdot10^{-3}$.
    \label{fig:results}}
\end{figure}

Figure~\ref{fig:results} (left) shows the likelihood as estimated by Annealed
Importance Sampling \cite{Salakhutdinov2009, Neal-2001} as a function of the
number of epochs
\footnote{While we do not report error margins for AIS likelihood estimates,
the numbers proved robust to changes in the number of particles and
temperatures being simulated. To obtain such robust estimates, we implemented
all the tricks described in \citet{Salakhutdinov2009} and
\cite{Salakhutdinov+Murray-2008}:
$p_A$ a zero-weight base-rate model whose biases are set by maximum likelihood;
interpolating distributions $p_i \propto p_A^{(1 - \beta_i)} p_B^{(\beta_i)}$,
with $p_B$ the target distribution; and finally analytical integration of all odd-layers.}.
Under this metric, MFNG achieves the fastest convergence, obtaining a
training/test set likelihood of $-71.26$/$-72.84$ nats after 94 epochs. In
comparison, MFNG-diag obtains $-73.22$/$-74.05$ nats and SML $-80.12$/$-79.71$
nats in 100 epochs. The picture changes however when plotting likelihood as a
function of CPU-time, as shown in Figure~\ref{fig:results} (right). Given a
wall-time of $8000$s for MFNG and SML, and $5000$s for MFNG-diag\footnote{This
discrepancy will be resolved in the next revision.}, SML is able to perform
upwards of $1550$ epochs, resulting in an impressive likelihood score of
$-64.94$ / $-67.73$. Note that these results were obtained on the
binary-version of MNIST (thresholded at 0.5) in order to compare to
\cite{Montavon2012arxiv}. These results are therefore not directly comparable to
\cite{Salakhutdinov2009}, which binarizes the dataset through sampling (by
treating each pixel activation as the probability $p$ of a Bernouilli
distribution).

\begin{figure}
  \centering
  \subfigure{
    \includegraphics[scale=0.38]{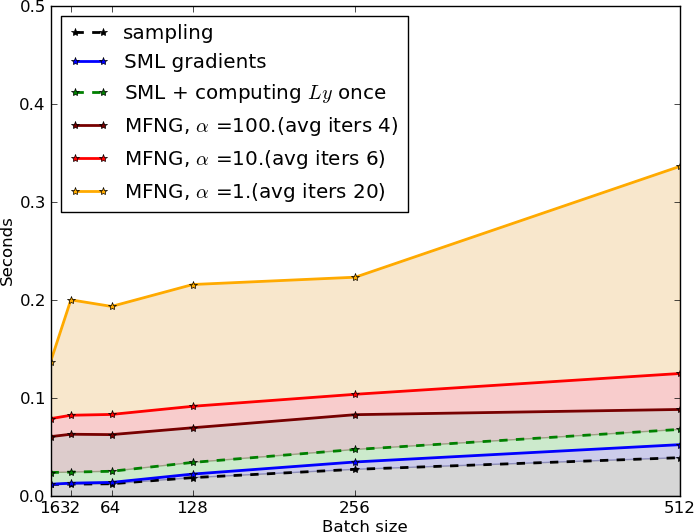}
  }
  \subfigure{
    \includegraphics[scale=0.38]{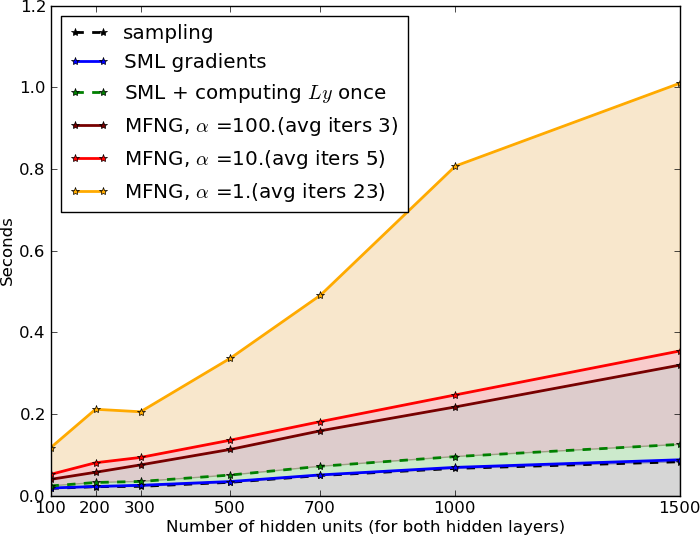}
  }
\caption[]{Breakdown of algorithm runtime, when we vary (left) the batch size
(with fixed model architecture $784-400-100$) and (right) the model size (with fixed batch size of
$256$). Runtime is additive in the order given by the labels (top to bottom).
Dotted lines denote intermediate steps, while continuous lines denote full
steps. Data was collected on a Nvidia GTX 480 card.\label{fig:timing}}
\end{figure}

Figure~\ref{fig:timing} shows a breakdown of the algorithm runtime, for various
components of the algorithm. These statistics were collected in the early stages
of training, but are generally representative of the bigger picture. While the
linear solver clearly dominates the runtime, there are a few interesting
observations to make. For small models and batch sizes greater than $256$,
a single evaluation of $Ly$ appears to be of the same order of
magnitude as a gradient evaluation. In all cases, this cost is smaller than
that of sampling, which represents a non-negligible part of the total
computational budget. This suggests that MFNG could become especially
attractive for models which are expensive to sample from. Overall however,
restricting the number of CG/MinRes iterations appears key to computational
performance, which can be achieved by increasing the damping factor $\alpha$.
How this affects convergence in terms of likelihood is left for future work.

\section{Discussion and Future Work}
\label{sec:discussion}

While the wall-clock performance of MFNG is not currently competitive with SML,
we believe there are still many avenues to explore to improve computational
efficiency. Firstly, we performed almost no optimization of the various MinRes
hyper-parameters. In particular, we ran the algorithm to convergence with a
fixed tolerance of $10^{-5}$. While this typically resulted in relatively few
iterations (around $15$), this level of precision might not be
required (especially given the stochastic nature of the algorithm).
Additionally, it could be worth exploiting the same strategy as HF where
the linear solver is initialized
by the solution found in the previous iteration. This may
prove much more efficient than the current approach of initializing the solver
with a zero vector. Pre-conditioning is also a well-known method for
accelerating the convergence speed of linear solvers \cite{Chapelle+Erhan-2011}.
Our implementation used a simple diagonal
regularization of $L$. The Jacobi preconditioner could be implemented easily
however by computing the diagonal of $L$ in a first-pass.



Finally, while our single experiment offers little evidence in support of either conclusion,
it may very well be possible that MFNG is simply not
computationally efficient for DBMs, compared to SML with centering. In this
case, it would be worth applying the method to either (i) models with known
ill-conditioning, such as factored 3-rd order Boltzmann Machines or (ii) models
and distributions exhibiting complex posterior distributions. In such
scenarios, we may wish to maximize the use of the positive phase statistics
(which were obtained at a high computational cost) by performing larger jumps
in parameter space. It remains to be seen how this would interact with SML,
where the burn-in period of the persistent chains is directly tied to the
magnitude of $\Delta\theta$.

\section*{Appendix}

We include the following derivations for completeness.

\subsection{Expected Hessian of $\log Z$ and Fisher Information.}

\begin{eqnarray*}
    \mathbb{E}_{p_\theta}\left[-\frac{\partial^{2}\log p(x)}{\partial\theta_{i}\partial\theta_{j}}\right] & = &
    \mathbb{E}_{p_\theta}\left[\frac{1}{p(x)^{2}}\frac{\partial p(x)}{\partial\theta_{j}}\frac{\partial p(x)}{\partial\theta_{i}}-\frac{1}{p(x)}\frac{\partial^{2}p(x)}{\partial\theta_{i}\partial\theta_{j}}\right]\\
    & = & \mathbb{E}_{p_\theta}\left[\left(\frac{1}{p(x)}\frac{\partial p(x)}{\partial\theta_{i}}\right)\left(\frac{1}{p(x)}\frac{\partial p(x)}{\partial\theta_{j}}\right)-\frac{1}{p(x)}\frac{\partial^{2}p(x)}{\partial\theta_{i}\partial\theta_{j}}\right]\\
    & = & \mathbb{E}_{p_\theta}\left[\frac{\partial\log p(x)}{\partial\theta_{i}}\frac{\partial\log p(x)}{\partial\theta_{j}}\right]-\sum_{x}p(x)\frac{1}{p(x)}\frac{\partial^{2}p(x)}{\partial\theta_{i}\partial\theta_{j}}\\
    & = & \mathbb{E}_{p_\theta}\left[\frac{\partial\log p(x)}{\partial\theta_{i}}\frac{\partial\log p(x)}{\partial\theta_{j}}\right]-\sum_{x}\frac{\partial^{2}p(x)}{\partial\theta_{i}\partial\theta_{j}}\\
    & = & \mathbb{E}_{p_\theta}\left[\frac{\partial\log p(x)}{\partial\theta_{i}}\frac{\partial\log p(x)}{\partial\theta_{j}}\right]-\frac{\partial^{2}\sum_{x}p(x)}{\partial\theta_{i}\partial\theta_{j}}\\
    & = & \mathbb{E}_{p_\theta}\left[\frac{\partial\log p(x)}{\partial\theta_{i}}\frac{\partial\log p(x)}{\partial\theta_{j}}\right]
\end{eqnarray*}

\subsection{Derivation of Equation~\ref{eq:dbm_metric}}

\begin{eqnarray*}
    \log p(x) & = & -E(x) - \log Z \\
\frac{\partial \log p(x)}{\partial\theta_i}
    & = & -\frac{\partial E(x)}{\partial \theta_i} - \frac{1}{Z}\sum_{x}\frac{\partial}{\partial\theta_i}\left[\exp\left(-E(x)\right)\right] \\
    & = & -\frac{\partial E(x)}{\partial \theta_i}
          + \mathbb{E}_{p_\theta} \left[ \frac{\partial E(x)}{\partial\theta_i} \right] \\
\mathbb{E_{p_\theta}}\left[-\frac{\partial^{2}\log p(x)}{\partial\theta_{i}\partial\theta_{j}}\right]
    & = &
        \mathbb{E}_{p_\theta}
        \left[
            \left(
                \frac{\partial E(x)}{\partial \theta_i}
                -\mathbb{E}_{p_\theta} \left[ \frac{\partial E(x)}{\partial\theta_i} \right]
            \right)
            \left(
                \frac{\partial E(x)}{\partial \theta_j}
                -\mathbb{E}_{p_\theta} \left[ \frac{\partial E(x)}{\partial\theta_j} \right]
            \right)
        \right] \\
    & = &
        \mathbb{E}_{p_\theta}
            \left[
                \frac{\partial E(x)}{\partial \theta_i}
                \frac{\partial E(x)}{\partial \theta_j}
            \right]
        - \mathbb{E}_{p_\theta} \left[ \frac{\partial E(x)}{\partial\theta_i} \right]
          \mathbb{E}_{p_\theta} \left[ \frac{\partial E(x)}{\partial\theta_j} \right]
\end{eqnarray*}

\small
\bibliography{ml,strings,strings-shorter}

\begin{thebibliography}{}

\bibitem[Amari(1985)Amari]{Amari-1985}
Amari, S. (1985).
\newblock Differential geometrical methods in statistics.
\newblock {\em Lecture notes in statistics\/}, {\bf 28}.

\bibitem[Amari(1998)Amari]{amari98natural}
Amari, S. (1998).
\newblock Natural gradient works efficiently in learning.
\newblock {\em Neural Computation\/}, {\bf 10}(2), 251--276.

\bibitem[Amari {\em et~al.}(1992)Amari, Kurata, and Nagaoka]{Amari-1992}
Amari, S., Kurata, K., and Nagaoka, H. (1992).
\newblock Information geometry of {B}oltzmann machines.
\newblock {\em {IEEE} Trans. on Neural Networks\/}, {\bf 3}, 260--271.

\bibitem[Carreira-Perpi{\~{n}}an and Hinton(2005)Carreira-Perpi{\~{n}}an and
  Hinton]{Perpinan+Hinton-2005-small}
Carreira-Perpi{\~{n}}an, M.~A. and Hinton, G.~E. (2005).
\newblock On contrastive divergence learning.
\newblock In {\em AISTATS'2005\/}, pages 33--40.

\bibitem[Chapelle and Erhan(2011)Chapelle and Erhan]{Chapelle+Erhan-2011}
Chapelle, O. and Erhan, D. (2011).
\newblock Improved preconditioner for hessian free optimization.
\newblock {\em NIPS Workshop on Deep Learning and Unsupervised Feature
  Learning\/}.

\bibitem[Freund and Haussler(1992)Freund and Haussler]{Freund+Haussler92}
Freund, Y. and Haussler, D. (1992).
\newblock A fast and exact learning rule for a restricted class of {Boltzmann}
  machines.
\newblock pages 912--919, Denver, CO. Morgan Kaufmann, San Mateo.

\bibitem[{LeCun} {\em et~al.}(1998){LeCun}, Bottou, Bengio, and
  Haffner]{LeCun98-small}
{LeCun}, Y., Bottou, L., Bengio, Y., and Haffner, P. (1998).
\newblock Gradient based learning applied to document recognition.
\newblock {\em Proc. IEEE\/}.

\bibitem[Martens(2010)Martens]{martens2010hessian}
Martens, J. (2010).
\newblock Deep learning via {H}essian-free optimization.
\newblock pages 735--742.

\bibitem[Montavon and Muller(2012)Montavon and Muller]{Montavon2012}
Montavon, G. and Muller, K.-R. (2012).
\newblock Deep {B}oltzmann machines and the centering trick.
\newblock In G.~Montavon, G.~Orr, and K.-R. Muller, editors, {\em Neural
  Networks: Tricks of the Trade\/}, volume 7700 of {\em Lecture Notes in
  Computer Science\/}, pages 621--637.

\bibitem[Montavon and M\"{u}ller(2012)Montavon and
  M\"{u}ller]{Montavon2012arxiv}
Montavon, G. and M\"{u}ller, K.-R. (2012).
\newblock Learning feature hierarchies with centered deep {B}oltzmann machines.
\newblock {\em CoRR\/}, {\bf abs/1203.4416}.

\bibitem[Neal(2001)Neal]{Neal-2001}
Neal, R.~M. (2001).
\newblock Annealed importance sampling.
\newblock {\em Statistics and Computing\/}, {\bf 11}(2), 125--139.

\bibitem[Paige and Saunders(1975)Paige and Saunders]{paige:1975}
Paige, C.~C. and Saunders, M.~A. (1975).
\newblock {Solution of Sparse Indefinite Systems of Linear Equations}.
\newblock {\em SIAM Journal on Numerical Analysis\/}, {\bf 12}(4), 617--629.

\bibitem[Pearlmutter(1994)Pearlmutter]{Pearlmutter-1994}
Pearlmutter, B. (1994).
\newblock Fast exact multiplication by the {H}essian.
\newblock {\em Neural Computation\/}, {\bf 6}(1), 147--160.

\bibitem[Raiko {\em et~al.}(2012)Raiko, Valpola, and LeCun]{Raiko-2012-small}
Raiko, T., Valpola, H., and LeCun, Y. (2012).
\newblock Deep learning made easier by linear transformations in perceptrons.
\newblock In {\em AISTATS'2012\/}.

\bibitem[Salakhutdinov and Hinton(2009)Salakhutdinov and
  Hinton]{Salakhutdinov2009}
Salakhutdinov, R. and Hinton, G. (2009).
\newblock Deep {B}oltzmann machines.
\newblock In {\em Proceedings of the Twelfth International Conference on
  Artificial Intelligence and Statistics (AISTATS 2009)\/}, volume~8.

\bibitem[Salakhutdinov and Murray(2008)Salakhutdinov and
  Murray]{Salakhutdinov+Murray-2008}
Salakhutdinov, R. and Murray, I. (2008).
\newblock On the quantitative analysis of deep belief networks.
\newblock volume~25, pages 872--879.

\bibitem[Schraudolph(1998)Schraudolph]{Schraudolph-1998}
Schraudolph, N.~N. (1998).
\newblock Centering neural network gradient factors.
\newblock In G.~B. Orr and K.-R. Muller, editors, {\em Neural Networks: Tricks
  of he Trade\/}, pages 548--548. Springer.

\bibitem[Tieleman(2008)Tieleman]{Tieleman08}
Tieleman, T. (2008).
\newblock Training restricted {B}oltzmann machines using approximations to the
  likelihood gradient.
\newblock pages 1064--1071.

\bibitem[Younes(1999)Younes]{Younes1999}
Younes, L. (1999).
\newblock On the convergence of {M}arkovian stochastic algorithms with rapidly
  decreasing ergodicity rates.
\newblock {\em Stochastics and Stochastic Reports\/}, {\bf 65}(3), 177--228.

\end{thebibliography}
\bibliographystyle{natbib}

\end{document}